\documentclass[sigconf]{acmart}

\def\BibTeX{{\rm B\kern-.05em{\sc i\kern-.025em b}\kern-.08emT\kern-.1667em\lower.7ex\hbox{E}\kern-.125emX}}
    
\copyrightyear{2019}
\acmYear{2019}
\setcopyright{rightsretained}
\acmConference[FAT* 2019]{ACM Conference on Fairness, Accountability, and Transparency}{February 2019}{Atlanta,GA, USA}
\acmDOI{}
\acmISBN{}

\begin{document}

%
\title{Tutorial: Safe and Reliable Machine Learning}

%
\author{Suchi Saria}
\affiliation{%
\department{Department of Computer Science}
  \institution{Johns Hopkins University}
  \city{Baltimore}
  \state{MD}
  \postcode{21218}
  \country{USA}}
\email{ssaria@cs.jhu.edu}

\author{Adarsh Subbaswamy}
\affiliation{%
\department{Department of Computer Science}
  \institution{Johns Hopkins University}
  \city{Baltimore}
  \state{MD}
  \postcode{21218}
  \country{USA}}
  \email{asubbaswamy@jhu.edu}

%
\begin{abstract}
This document serves as a brief overview of the ``Safe and Reliable Machine Learning'' tutorial given at the 2019 ACM Conference on Fairness, Accountability, and Transparency (FAT* 2019). The talk slides can be found here: \url{https://bit.ly/2Gfsukp}, while a video of the talk is available here: \url{https://youtu.be/FGLOCkC4KmE}, and a complete list of references for the tutorial here: \url{https://bit.ly/2GdLPme}.
\end{abstract}

\maketitle


\section{Motivation and Outline}
Machine Learning driven decision-making systems are starting to permeate modern society---for example, to decide bank loans, criminals' incarceration, clinical decision-making, and the hiring of new employees. As we march towards a future where these systems underpin most of society's decision-making infrastructure, it is critical for us to understand the principles that will help us engineer for reliability. In this tutorial, we (1) give an overview of issues to consider when designing for reliability, (2) draw connections to concepts of fairness, transparency, and interpretability, and (3) discuss novel technical approaches for measuring and ensuring reliability.

\section{Principles of Reliability}
The field of machine learning, despite its increasing use in high-stakes and safety-critical applications, fundamentally lacks a framework for reasoning about failures and their potentially catastrophic effects. This is in contrast to traditional engineering disciplines which have been forced to consider the safety implications across a broad set of applications, from building a bridge to managing a nuclear power plant. Bridging across these applications is the discipline of \emph{reliability engineering} (see, e.g., \citep{reliability}) which seeks to ensure that a product or system performs as intended (without failure and within specified performance limits). Drawing on this notion of reliability, we have pulled out three principles of reliability engineering that we use to group and guide technical solutions for addressing and ensuring reliability in machine learning systems:
\begin{enumerate}
    \item \textbf{Failure Prevention:} Prevent or reduce the likelihood of failures.
    \item \textbf{Failure Identification \& Reliability Monitoring:} Identify failures and their causes when they occur.
    \item \textbf{Maintenance:} Fix or address the failures when they occur.
\end{enumerate}

In what follows we will consider each of the principles of reliability in turn, summarizing key approaches when they exist and speculating about open problem areas. The focus of this tutorial is on supervised learning (i.e., classification and regression). For an overview of issues associated with reinforcement learnings see \citep{amodei2016concrete}.

\section{Failure Prevention}
To prevent failures, ideally we could \emph{proactively} identify likely sources of error and develop methods that correct for these in advance. This requires us to explicitly reason about common sources of errors and issues. We broadly categorize four sources of failures and discuss them each: 1) bad or inadequate data, 2) differences or shifts in environment, 3) model associated errors, and 4) poor reporting.

\subsection{Bad or Inadequate Data}
Inadequate data can cause errors related to differential performance. For example, when a particular class or subpopulation is underrepresented in a dataset, the performance of a classifier on these subgroups can be very poor even though average or overall accuracy is high (e.g., \citep{buolamwini2018gender}). These errors can be avoided by measuring performance on subpopulations of interest. If key subpopulations have not been identified, then one could consider clustering data to find regions of poor support. Inadequate data can be addressed by collecting more representative data and through better design of of the objective function.

On the other hand, bad data refers to cases in which the data simply do not contain the information necessary to answer the question or perform the task of interest. Understanding when machine learning can be applied is crucial to avoiding model misuse. Some examples are discussed in the tutorial, such as trying to predict behavioral traits from facial images.

\subsection{Differences or Shifts in Environment}
Differences between training and deployment environments can lead to degraded model performance and failures post-deployment. As an example, in \citep{zech2018variable} the authors train a model to diagnose pneumonia from chest X-rays at a particular hospital. When evaluated on that dataset, the model yielded good performance. But when evaluated at two other hospital networks the performance was significantly worse, calling into question the generalizability or \emph{external validity} of the model.

The issue is that modelers typically assume that training data is representative of the target population or environment where the model will be deployed. Yet commonly there is bias specific to the training dataset which causes learned models to be unreliable: they do not generalize beyond the training population and, more subtly, are not robust to shifts in practice patterns or policy in the training environment. This bias can arise due to the method of data collection, frequently due to some form of selection bias. The bias may also be caused by differences between the policy or population in the training data and that of the deployment environment. In some instances, the very deployment of the decision support tool can change practice and lead to future shifts in policy in the training environment, which leads to dangerous \emph{feedback loops} (consider the example of predictive policing algorithms \citep{lum2016predict}).

The machine learning community has extensively studied this problem of \emph{dataset shift} in which training and test distributions are different \citep{candela2009dataset}. However, the solutions have primarily been \emph{reactive} in that they use (unlabeled) samples from the target distribution in combination with training data during learning to optimize directly for the target environment. But, in general, it is not feasible to access data from all possible test environments at training time (in fact, the deployment environment may be unspecified or may exist in the future). Thus, we instead want a predictive model that generalizes to new, unseen environments. Achieving this requires a shift in paradigm to \emph{proactive} approaches \citep{subbaswamycounterfactual,subbaswamy2019transport}: in order to prevent failures we should learn a model that is explicitly protected against problematic shifts that are likely to occur.

In this subsection of the tutorial we go through the framework laid out in \citep{subbaswamy2019transport} for preventing failures due to shifts in environment. The framework is enticing because it allows model developers to proactively reason about the possible shifts that could occur in their application, and then dictates what they need to model in order to make optimal predictions that are unaffected by the shifts they want to ignore. An overview is as follows: The framework uses directed acyclic graphs (DAGs) as a language for representing shifts in the underlying data generating process (DGP) that could occur between environments. This language is sufficiently expressive to include common forms of dataset shift (such as covariate and label shift) as well as more complex ones including policy shift (which leads to feedback loops \citep{schulam2017reliable}). The graphical representation of a DGP can be augmented (into a \emph{selection diagram} \citep{pearl2011transportability}) to identify the shifts we want to guard against. Thus, the graph serves as an invariance specification that a reliable model needs to satisfy. The algorithm presented in \citep{subbaswamy2019transport} serves as a preprocessing step that dictates which pieces of the DGP to model, and subsequently these pieces can be fit using arbitrarily complex models (e.g., neural networks) and combined to make predictions that satisfy the invariance requirements. For a discussion of other properties of this procedure, see the tutorial video and the paper \citep{subbaswamy2019transport}.

\subsection{Model Associated Errors}
Broadly, we can think of two main types errors related to modeling choices:
\begin{enumerate}
    \item Faulty (implicit) model assumptions
    \item Fragile models
\end{enumerate}

Faulty model assumptions are often due to \emph{model misspecification}. For example, a linear model may be inappropriate to use in a complex setting with many interactions (i.e., nonlinearities) between the features. To reduce or prevent model misspecification bias, we should make meaningful use of inductive bias in our choice of learner or alternatively consider nonparametric methods. Another example of a faulty model assumption is the case of \emph{dependent data}. It is common practice to assume that samples have independent errors, but this assumption does not hold when working with geographic data or social network data where outcomes are tied. The lesson here is to be explicit about any assumptions that are made (which is related to having good reporting practices (Section 3.4)).

The issue with \emph{model fragility} occurs when models are applied to problems with high dimensional inputs. The model is ``fragile'' in the sense that its predictions are very sensitive to small perturbations of the input. An example of this which has received attention in recent years is the problem of \emph{adversarial examples} (e.g., \citep{goodfellow2014explaining}). The high dimensional inputs are images, and by perturbing the images with human-imperceptible noise the model prediction can change to be confidently wrong. This sort of failure is complementary to dataset shift because despite proactive correction, a model can be susceptible to small perturbations as a result of its parameterization (i.e., it is a fitting issue).

Research into adversarial training has produced methods that have useful properties for ensuring reliability. For example, these methods consider \emph{robust objectives} in which the goal is to minimize the loss achieved by the worst-case adversarial attack. Further, some training methods produce \emph{certificates of robustness} (e.g., \citep{sinha2018certifying}) which give data-dependent bounds on the worst-case loss. Similarly, methods for model \emph{verification} provide a means for making yes/no statements about individual input-output pairs which let us ``stress test'' a model (e.g., \citep{dvijotham2018dual}).

\subsection{Poor Reporting}
An important source of error is the mismatch between a model's intended purpose and the way it is actually used, which is often a result of poor reporting practices. As opposed to other high-impact industries (such as transportation or pharmaceuticals), few standards exist for reporting and documentation in machine learning. A number of recent proposals seek to address this by creating templates for appropriate documentation, including ``datasheets for datasets'' \citep{gebru2018datasheets} and ``model cards for model reporting'' \citep{mitchell2019model}. These proposals advocate explicitly stating intended use, details of creation, ethical considerations, and many other relevant factors. We think a natural extension is to also include reliability criteria in model reporting, since guarantees are a promise regarding reliable behavior that should be documented. For example, statements about likely shifts in environment that were considered, certificates of robustness, and model verification would improve reporting practices. The primary open question regarding reporting is: what are the relevant aspects that should be required in documentation, and who should decide and enforce such requirements?

\subsection{Relating the Sources of Failures}
\begin{figure}[!th]
        \center{\includegraphics[width=0.45\textwidth]
        {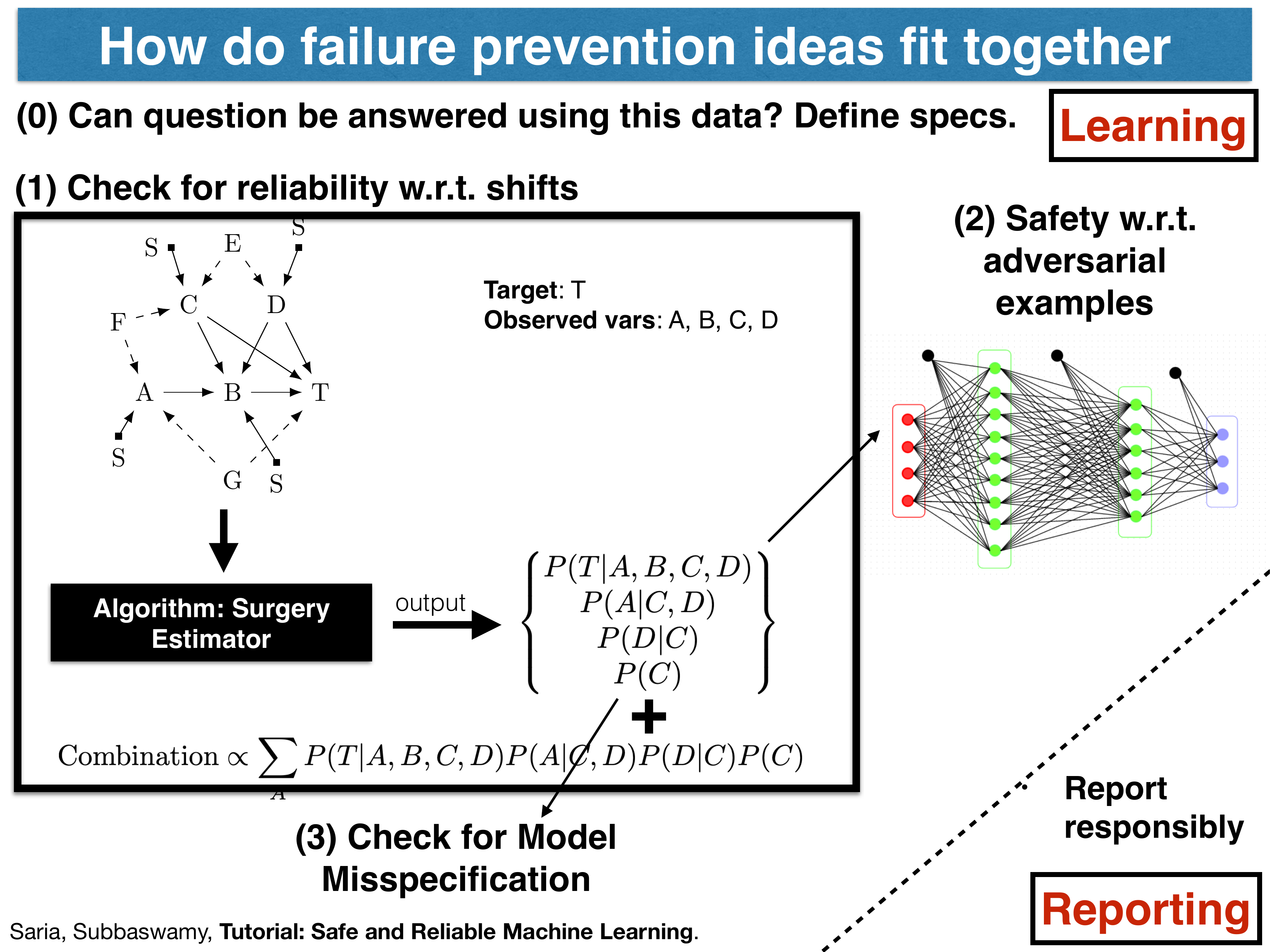}}
        \caption{\label{fig:put-together} Diagram showing how the various sources of errors relate to reliability.}
\end{figure}
The various sources of errors and the corresponding proactive methods for preventing them can be integrated together into a checklist for ensuring reliability and failure-proofing the development process, as shown in Fig \ref{fig:put-together}. We should first ask: can the question of interest be answered using the available data? Then, we determine reliability with respect to shifts, check for safety regarding adversarial examples, and we can check for model misspecification. Finally, we should make sure we have responsibly documented the model and training procedure. As discussed in the tutorial, there are open problems related to each source of failure, and addressing them will be critical for the success of reliable machine learning.

\section{Failure Identification and Reliability Monitoring}
Failure prevention takes place prior to and during model learning and development. Once the system has been deployed we require a means for \emph{test-time monitoring}. A useful approach is to assess \emph{point-wise reliability}: assess the model output for each new input, rejecting the output when it is deemed unreliable. This is closely related to the concept of detecting ``out-of-distribution'' examples: samples generated by a process that is different than the process generating the ``nominal'' data. These examples are, in some sense, ``far away'' from the known data distribution. Extensive literature for this exists under the names \emph{anomaly detection} and \emph{open-category detection} (see full reference link for more information).

In this section of the tutorial we consider how to compute point-wise ``trust score'' to audit a model's prediction to determine whether or not to reject the prediction \citep{jiang2018trust,schulam2019rue}. We primarily consider two criteria for performing the audit: the \emph{density principle} and the \emph{local fit principle}. The density principle essentially asks if the test case is close to training samples, while the local fit principle asks if the model was accurate on training samples close to the test case. By synthesizing these two it is possible to audit model predictions subsequent to model training (see \citep{schulam2019rue} for details).

While this section is primarily focused on purely algorithmic approaches to test-time monitoring, it is worth mentioning that there has been work on human-driven monitoring. These approaches include \emph{crowdsourcing} and \emph{human-in-the-loop debugging}, and are particularly useful because they are often effective at identifying a model's ``unknown unknowns'' (predictions for which a model is confident despite being wrong) \citep{attenberg2011beat}. For some sample references consult the relevant section of the tutorial slides.

\section{Maintenance}
Model maintenance remains a largely open area for ensuring reliability. For example, beyond error monitoring is the question of how to detect when updates to the model are necessary. Further, how do we safely update the model? There may be issues of forgetting what was previously learned or feedback loops created during continuous learning. The challenges of maintenance are also compounded by the fact that AI systems are complex with multiple components. See \citep{sculley2014machine} for a lucid discussion of the maintenance costs associated with the ``technical debt'' incurred by machine learning development.

\section{Conclusion}
In this tutorial we have defined reliability as a vital property of a successful machine learning system: given a specification of desired behavior, we want to ensure that the machine learning system behaves consistently as intended and is not error prone. We note that there is some existing work in this direction under the name ``robust machine learning.'' However, we believe that robustness, while an important aspect of reliability (e.g., in guarding against adversarial examples), is too narrow in scope (in part due to its connotations with the well-established field of robust statistics) and fails to address important sources of failure such as model misuse due to poor reporting. We also note that reliability is separate from other desirable properties such as privacy, fairness, and transparency. We identified three principles for ensuring reliability: failure prevention, failure identification and reliability monitoring, and maintenance. It is our hope that these principles will guide future work and discussions about successfully deploying machine learning across a variety of domains.

\bibliographystyle{ACM-Reference-Format}
\bibliography{refs}
\end{document}